\documentclass[letterpaper, 10 pt, conference]{ieeeconf}  

\IEEEoverridecommandlockouts                              

\makeatletter
\let\NAT@parse\undefined
\makeatother

\usepackage[comma,numbers, compress]{natbib}

\usepackage[hidelinks]{hyperref}
\usepackage{url}
\usepackage{blindtext}
\usepackage{graphicx}
\usepackage[usenames,dvipsnames]{xcolor}
\usepackage[draft]{fixme}
\fxsetup{theme=color}

\usepackage{amsmath}
\usepackage{amssymb}

\usepackage{booktabs}
\usepackage{textcomp}
\usepackage{threeparttable}

\usepackage[capitalize]{cleveref}

\usepackage{tikz}
\usetikzlibrary{arrows}
\usetikzlibrary{positioning,calc}
\usetikzlibrary{decorations.pathreplacing}
\usetikzlibrary{decorations.markings}
\usetikzlibrary{fit}
\usetikzlibrary{shapes.callouts}
\usetikzlibrary{shapes.geometric}
\usetikzlibrary{matrix}

\usepackage[firstpage=true]{background}

\newcommand\copyrighttext{%
        \parbox{\textwidth}{
                \footnotesize
                In Proceedings of IEEE/RSJ International Conference on Intelligent Robots and Systems (IROS), Vancouver, Canada, September 2017, \\DOI: 10.1109/IROS.2017.8206310 
        }
}

\SetBgContents{\copyrighttext}
\SetBgScale{1}
\SetBgColor{black}
\SetBgAngle{0}
\SetBgOpacity{1}
\SetBgPosition{current page.north}
\SetBgVshift{-0.8cm}

\graphicspath{{figures/}}
\title{\LARGE \bf
Anytime Hybrid Driving-Stepping Locomotion Planning
}

\author{Tobias Klamt and Sven Behnke
\thanks{All authors are with Rheinische Friedrich-Wilhelms-Universit\"at Bonn, Computer Science Institute VI, 
		Autonomous Intelligent Systems, Friedrich-Ebert-Allee 144, 53113 Bonn, Germany
        {\tt\small klamt@ais.uni-bonn.de, behnke@cs.uni-bonn.de}. This work was supported by the European Union's Horizon 2020 Programme under 
        Grant Agreement 644839 (CENTAURO).}%
}

\begin{document}

\maketitle
\thispagestyle{empty}
\pagestyle{empty}

\begin{abstract}

Hybrid driving-stepping locomotion is an effective approach for navigating in a variety of environments. Long, sufficiently even distances can be quickly covered by driving while obstacles can be overcome by stepping. Our quadruped robot Momaro, with steerable pairs of wheels located at the end of each of its compliant legs, allows such locomotion. Planning respective paths attracted only little attention so far. 

We propose a navigation planning method which generates hybrid locomotion paths. The planner chooses driving mode whenever possible and takes into account the detailed robot footprint. If steps are required, the planner includes those. To accelerate planning, steps are planned first as abstract manoeuvres and are expanded afterwards into detailed motion sequences. Our method ensures at all times that the robot stays stable. Experiments show that the proposed planner is capable of providing paths in feasible time, even for challenging terrain. 

\end{abstract}


\section{Introduction}

Hybrid driving-stepping locomotion enables robots to traverse a wide variety of terrain types. Application domains, such as search and rescue and delivery services, pose considerable navigation challenges for robots due to non-flat grounds. Sufficiently flat terrain can be traversed by driving, which is fast, efficient and safe, regarding the robot stability. However, driving traversability is limited to moderate slopes and height differences and obstacle-free paths. Stepping locomotion requires only adequate footholds and, hence, enables mobility in cases where driving is unfeasible. But stepping is also slower and decreases the robot stability.

\begin{figure}
\centering
\includegraphics[width=0.90\linewidth]{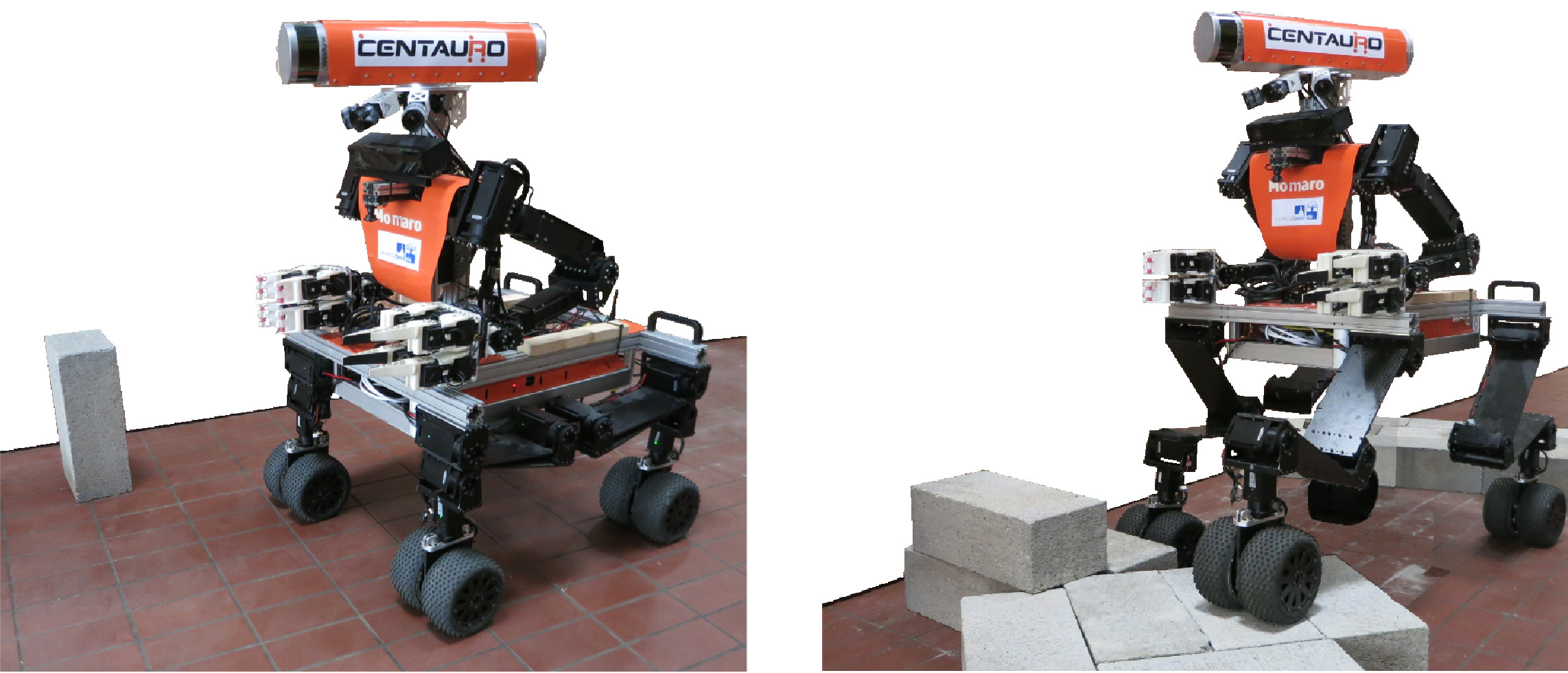}
\caption{Our hybrid wheeled-legged mobile manipulation robot Momaro is capable of omnidirectional driving (left) and stepping (right).}
\label{fig:momaro}
\end{figure}

Most mobile ground robots use either driving locomotion or stepping locomotion, and there exist path planning methods for both such locomotion modes independently~\cite{colas20133d,menna2014real,brunner2012motion,ziaei2014global,kalakrishnan2011learning,wermelinger2016navigation,perrin2016continuous}. Our mobile manipulation robot Momaro~\cite{Schwarz:ICRA2016} (see \cref{fig:momaro}), however, supports both locomotion types due to its four legs ending in steerable pairs of wheels. This unique design allows omnidirectional driving on sufficiently flat terrain and stepping to overcome obstacles. In contrast to purely walking robots, Momaro is able to change its configuration of ground contact points (which we will refer to as its footprint) under load without lifting a foot. This enables motion sequences for stepping that have large stability margins.
Multiple platforms that are capable of driving-stepping locomotion have been developed~\cite{halme2003workpartner, takahaashi2006rough, bae2016walking, stentz2015chimp, hashimoto2005realization}, but planning which combines the advantages of both locomotion types was addressed for none of these.

In our previous work with Momaro~\cite{frontiers2016}, we demonstrated semi-autonomous driving, 2D $(x,y)$ path planning and execution in a Mars-like environment accompanied by manipulation tasks. In this work, we extend the driving path planning method to incorporate the robot orientation $\theta$ and its detailed footprint in order to increase driving flexibility. We improve the path quality by introducing an orientation cost term.

In addition, we demonstrated stepping over a wooden bar obstacle, climbing stairs, and egressing a car with Momaro at the DARPA Robotics Challenge (DRC)~\cite{nimbro_rescue}. All of the DRC tasks were performed via teleoperation based on pre-defined motion sequences. Teleoperation depends on a good data connection between the operator station and robot and generates a high cognitive load for the operators. Autonomous locomotion is desirable to relieve the operators and to increase speed and safety.

We extend the locomotion planner to generate stepping motions. Driving in difficult terrain and stepping require a high planning resolution which increases planning times. To keep the search space feasible, we first generate abstract steps that we later expand to detailed motion sequences. 

\noindent To summarize, the main contributions in this paper are:    
\begin{itemize}
	\item a three-dimensional $(x,y,\theta)$  driving path planning method allowing driving in constrained uneven environments by consideration of the detailed robot footprint,   
	\item the introduction of orientation costs, favoring a preferred driving direction to align the robot with the path,
	\item a hierarchical step planner, which generates detailed manoeuvres to perform individual steps under the constraint to always keep the robot statically stable, and
	\item application of anytime planning to quickly find paths with bounded suboptimality. 
\end{itemize}

We demonstrate our approach in simulation and with the real robot and systematically evaluate the effect of our acceleration methods. The results indicate that our planner provides paths in feasible time even for challenging tasks.


\section{Related Work}

Path planning in unstructured terrain has been addressed by many works. The considered systems provide either purely wheeled/tracked locomotion or are able to traverse terrain by walking. Planning is often done with either grid-based searches, such as A*~\cite{hart1968formal}, or sampling-based approaches, such as RRT~\cite{lavalle1998rapidly}. Despite the application of similar planning methods, these two locomotion modes differ in many aspects. 

Driving is fast and energy efficient on sufficiently flat terrain, which makes it suitable for traversing longer distances. When supported by three or more wheels, the robot is generally statically stable. Planning of drivable paths in unstructured environments is heavily dependent on the degrees of freedom (DoF) of the platform. Simple robot designs offer longitudinal and rotational movements with a constant robot shape~\cite{gerkey2008planning},~\cite{howard2007optimal}. For search and rescue scenarios, some robots were extended by tracked flippers~\cite{colas20133d},~\cite{menna2014real},~\cite{brunner2012motion}. These allow the robots to climb stairs and thus increase capabilities but also planning complexity due to additional shape shifting DoFs. Flipper positions are often not considered by the initial navigation path planning and are adjusted to the terrain in a second planning step. Platforms which offer omnidirectional locomotion increase the path planning search space by another dimension~\cite{ziaei2014global}. Driving is restricted, however, by terrain characteristics such as height differences and slopes which makes it not suitable for very rough terrain and for overcoming obstacles. 

Legged locomotion is capable of traversing more difficult terrain since it only requires isolated feasible footholds. The drawback of this locomotion mode is that motion planning is much more complex. Since legs are lifted from the ground repeatedly, the robot also has to constantly ensure that it remains stable. Due to the high motion complexity of stepping, path planning is often performed in at least two hierarchical levels~\cite{kalakrishnan2011learning},~\cite{wermelinger2016navigation},~\cite{perrin2016continuous}. A coarse planning algorithm identifies feasible footholds or areas for feasible steps. Detailed motion planning is done in a second step to connect these footholds. Navigation towards the goal is either included in the coarse planning or realized in a higher-level planner.

Since both locomotion modes have complementary advantages, it is promising to combine those. Halme et al.~\cite{halme2003workpartner} and Takahaashi et al.~\cite{takahaashi2006rough} developed quadruped robots with legs ending in wheels. Control mechanisms to overcome obstacles are presented, but locomotion planning is not addressed. The hybrid locomotion robots HUBO~\cite{bae2016walking} and CHIMP~\cite{stentz2015chimp} were used by the winning and the third best teams at the DARPA Robotics Challenge. HUBO provides legged and wheeled locomotion, but needs to shift its shape to switch between those. Thus, hybrid locomotion, which combines advantages of both locomotion types, is not possible. CHIMP provides bipedal and quadruped hybrid locomotion. However, hybrid locomotion planning is not presented. Finally, a bipedal robot, capable of driving and walking, and a respective planning algorithm is presented by Hashimoto et al.~\cite{hashimoto2005realization}. Depending on the terrain, it either chooses walking or driving mode. A combination of both, which might bring further advantages, is not considered. Recently, Boston Dynamics introduced its biped platform Handle\footnote{\url{https://youtu.be/-7xvqQeoA8c}} with legs ending in wheels. It demonstrated manoeuvres which require very good dynamic control but path planning was not presented.

Our approach combines both locomotion modes in a single planning algorithm and thus has many benefits of both.


\section{Hardware}

We use our quadruped robot platform Momaro~\cite{Schwarz:ICRA2016} (see~\cref{fig:momaro}) with articulated legs ending in directly-driven 360\textdegree~steerable pairs of wheels. Those offer omnidirectional driving and the possibility to change the robot footprint under load which neither can be done by pure driving nor by pure walking robots and enables novel movement strategies. 

Each leg consists of three pitch joints which allow leg movements in the sagittal plane. Lateral leg movements are possible only passively. Legs show compliant behaviour due to their elastic carbon composite links, which work as a passive suspension system on rough terrain. Moreover, soft foam-filled wheels compensate small terrain irregularities. 

A continuously rotating Velodyne Puck 3D laser scanner with spherical field-of-view at the robot head and an IMU provide measurements for terrain perception.


\section{Environment Representation}

\begin{figure}
\centering
\includegraphics[width= .85\linewidth]{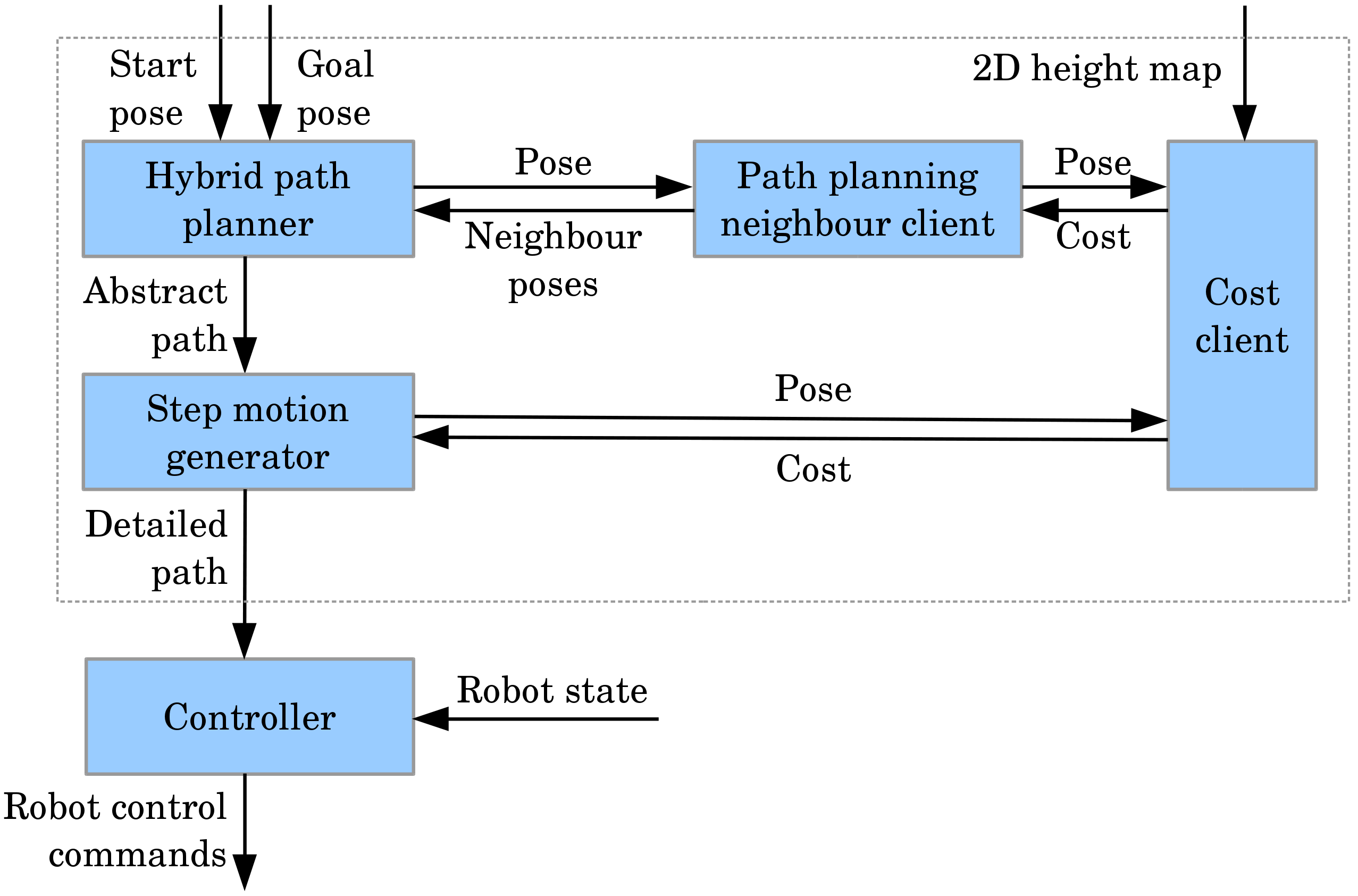}
\caption{Hybrid locomotion system structure: The hybrid path planner searches an abstract path from start to goal pose. The step motion generator expands this abstract path to a detailed path which can be executed by the robot. A neighbour client provides neighbour states to the planner. Both, this neighbour client and the step motion generator, request pose costs from the cost client which generates costs out of the 2D height map. The resulting path is executed by a controller.}
\label{fig:system_overview}
\end{figure}

An overview of the planning system structure is given in \cref{fig:system_overview}. Range measurements from the 3D laser scanner are used for mapping and localization by utilizing a multiresolution surfel map~\cite{droeschel2017continuous}. Input to the planner are a 2D height map and the start and goal robot poses. A robot pose \mbox{$\vec{r}=(r_x,r_y,r_\theta, \vec{f_1}, ... ,\vec{f_4})$} includes the robot base position $r_x, r_y$ and orientation $r_\theta$ and the foot positions $\vec{f}_j = (f_{j,x}, f_{j,y})$ in map coordinates. The grid resolution is set to 2.5\,cm with 64 possible orientations at each position. 

\begin{figure}
\centering
\includegraphics[width=0.9\linewidth]{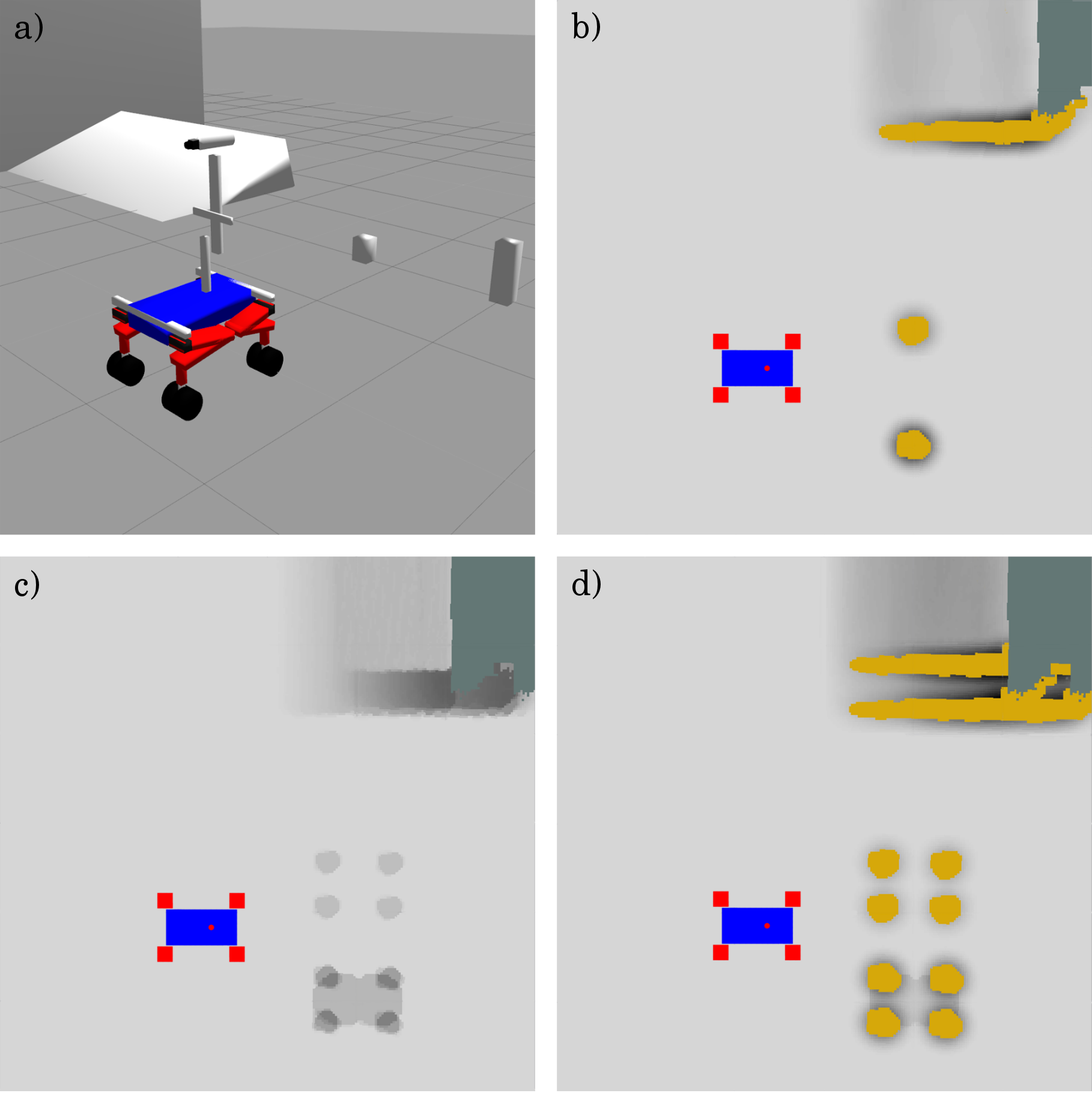}
\caption{Driving cost computation: a) Simulated scenario in which the robot stands in front of a ramp, a small and a tall pole. b) Foot costs. Yellow areas are not traversable by driving, olive areas are unknown. \mbox{c) Body costs. The robot can take the small} pole between its legs while the tall pole generates costs for lifting the robot body. Costs are shown for the current robot orientation. d) Pose costs combine body costs and foot costs at their respective positions.}
\label{fig:cost}
\end{figure}

The cost client computes pose cost values from the height map for a robot pose as follows: From the height map, local unsigned height differences $\bigtriangleup H$ are computed to generate the foot specific cost
\begin{equation}
C_\text{F}(c_j) = 1 + k_1 \cdot \sum\nolimits_{c_\text{i} \in \text{map}} \bigtriangleup H(c_\text{i}) \cdot w(c_\text{i})
\end{equation}
where $k_1$ = 100 and
\begin{equation}
w(c_i) = 
	\begin{cases}
			\infty & \text{if}~\left\lVert c_\text{i} - c_j\right\rVert \textless r_\text{F} \land \bigtriangleup H(c_\text{i}) \textgreater 0.05, \\
			1- \frac{\left\lVert c_\text{i} - c_j\right\rVert}{r_N} & \text{if}~\left\lVert c_\text{i} - c_j\right\rVert \textless r_\text{N}, \\
			0 & \text{otherwise}
	\end{cases}
\end{equation}
for a map cell $c_j$ in which the foot $\vec{f}_j$ is located. Foot costs are assigned an infinite value if untraversable height differences \textgreater ~0.05\,m occur in a surrounding of the size of a foot ($r_\text{F}$ = 0.12\,m). In a neighbourhood of greater size ($r_\text{N}$ = 0.3\,m), height differences are accumulated weighted by their distance to $c_j$. Foot costs are defined to be 1 in flat surroundings and increase if challenging terrain occurs. $C_\text{F}$ includes traversability information and describes the surrounding of each foot position (see~\cref{fig:cost}). 

The robot shape allows obstacles to pass between the robot legs. However, if obstacles are too high they might collide with the robot base. The base cost 
\begin{equation}
	C_B(\vec{r}) = 1 + k_2 \cdot \max(H_{\text{max,uB}} - H_\text{B}, 0) + k_3 \cdot \bigtriangleup H_{\text{max,F}} \text{,}
\end{equation}
where $k_2$ = 1 and $k_3$ = 0.5 compares the maximum terrain height under the robot body $H_{\text{max,uB}}$ with the body height $H_\text{B}$ and assigns additional costs if the space is not sufficient. In addition, the height difference between the lowest and highest foot $\bigtriangleup H_\text{max,F}$ generates costs since this is a measure for the terrain slope under the robot. Again, the basic cost is 1 which increases for challenging terrain. The robot base is estimated by two circles of 0.25\,m radius to avoid expensive detailed collision checking.

All cost values are combined into the pose cost  
\begin{equation}
	C(\vec{r}) = k_4 \cdot \max_{j}(C_{\text{F}}(\vec{f}_j)) + k_5 \cdot \sum_{j=1}^{4} C_\text{F}(\vec{f}_j) + k_6 \cdot C_B(\vec{r}) \text{,}
\end{equation}
where $k_4$ = 0.1, $k_5$ = 0.1 and $k_6$ = 0.5. Pose costs are defined to be 1 on flat terrain where both, foot and body costs, induce 50\% of the pose costs. We want to consider the terrain under all four wheels but want to prefer a pose with four slightly challenging contact points over a pose with three non-challenging and one very difficult contact point. Hence, it is neither sufficient to sum up all individual foot costs nor to just take the maximum. A weighted sum of both, however, achieves the desired functionality.


\section{Path Planning}

\begin{figure}
\centering
\includegraphics[width=0.55\linewidth]{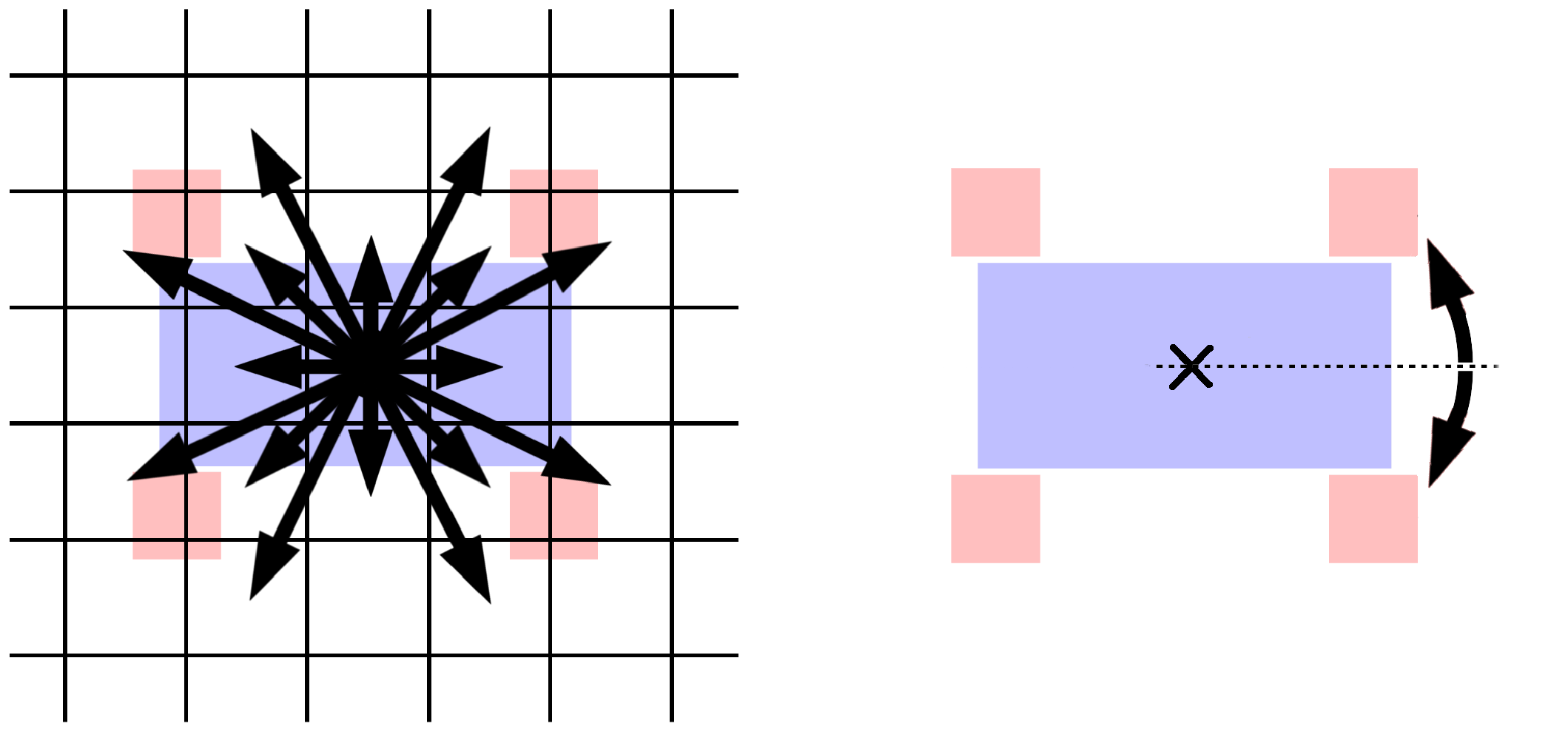}
\caption{Driving locomotion neighbour states can either be found by straight moves with fixed orientation within a 16-neighbourhood (l.) or by orientation changes on a fixed position (r.). Grid and orientation resolution are enlarged for better visualization.}
\label{fig:driving_neighbours}
\end{figure}

Path planning is done in a hybrid planner, which prefers the driving mode and considers steps only if necessary. It is realized through an A*-search on a pose grid. The used heuristic combines the Euclidean distance with orientation differences. For each pose, the path planning neighbour client provides feasible neighbouring poses (see~\cref{fig:system_overview}). Driving neighbours can be found within a 16-neighbourhood and by turning on the spot to the next discrete orientation~(\cref{fig:driving_neighbours}).

\begin{figure}
\centering
\includegraphics[width=0.9\linewidth]{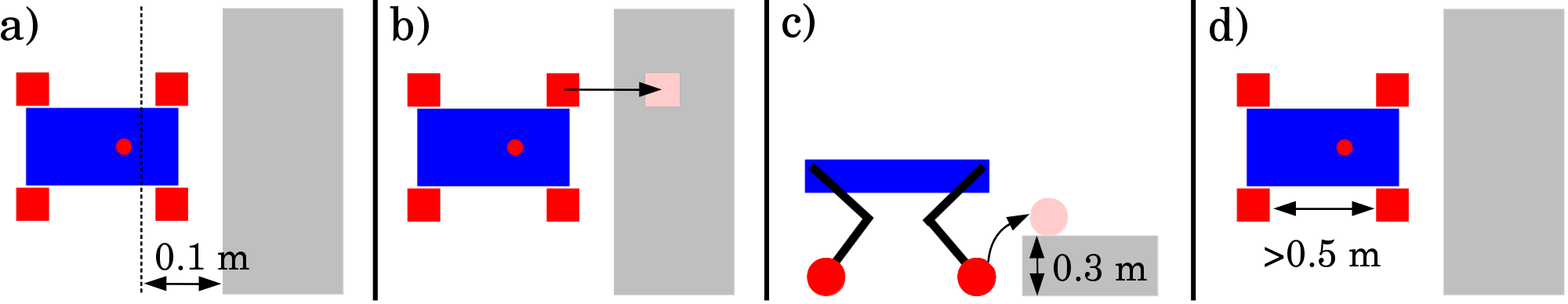}
\caption{Criteria to add steps to the hybrid planner: a) A foot is close to an obstacle, b) a feasible foothold can be found, c) the height difference to overcome is $\leq$ 0.3 m and d) the distance between the feet on the ``non-stepping'' robot side is \textgreater \,0.5\,m. Grey areas show an elevated platform.}
\label{fig:step_criteria}
\end{figure}

As illustrated in~\cref{fig:step_criteria}, additional stepping manoeuvres are added, if a foot $\vec{f}_j$ is
\begin{itemize}
	\item close to an obstacle 
	
	$\left(\exists c\in map~ \left( C_\text{F}(c)=\infty \land \left\lVert c - \vec{f}_j\right\rVert < 0.1~\text{m}\right)\right) $,
	\item a feasible foothold $c_\text{h}$ with $C_\text{F}(c_\text{h})$ \textless $\infty$ can be found in front of the foot in its sagittal plane that respects a maximum leg length,
	\item the height difference to the foothold is small \mbox{($\lvert H(\vec{f}_j) - H(c_\text{h})\rvert \leq 0.3\,\text{m}$)}, and
	\item the distance between the two feet on the ``non-stepping'' robot side is \textgreater \,0.5\,m to guarantee a safe stand while stepping.
\end{itemize}
A step is represented as an additional possible neighbouring pose for the planner.

\begin{figure}
\centering
\includegraphics[width=0.85\linewidth]{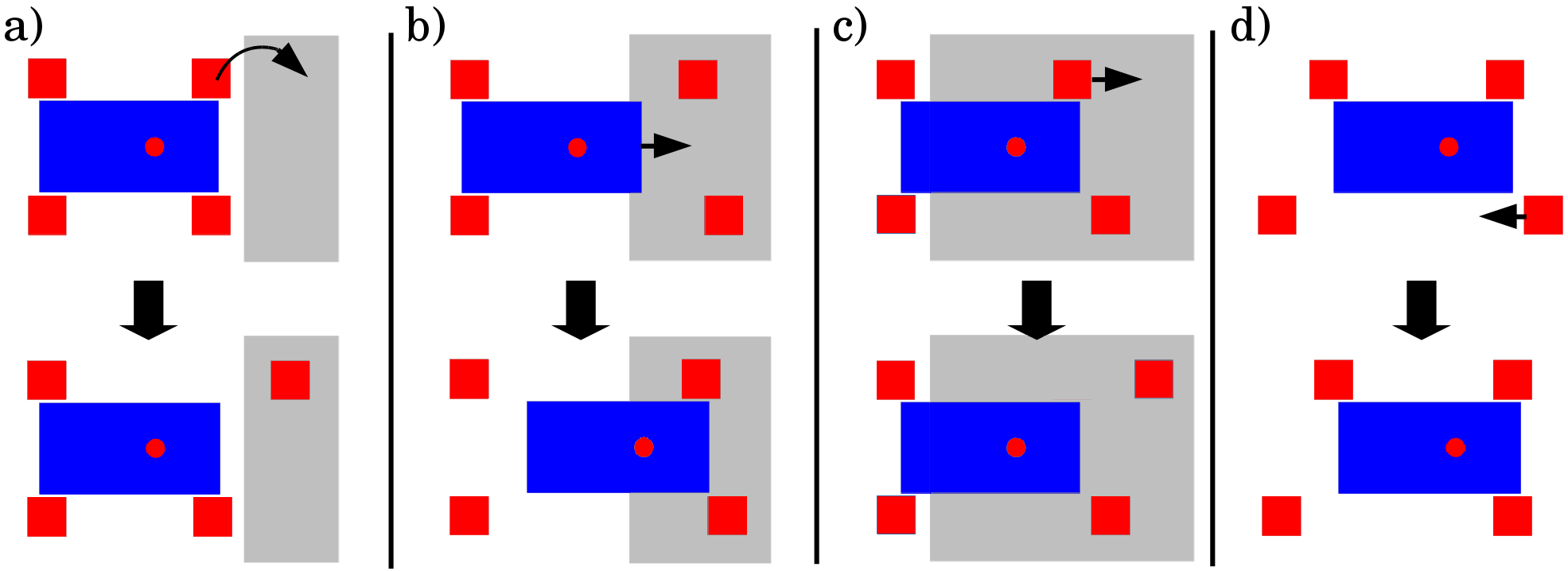}
\caption{Manoeuvres which extend the driving planner to a hybrid locomotion planner, visualized on a height map: a) Abstract steps, b) longitudinal base shifts, c) driving a pair of wheels at one front foot forward, and d) driving a pair of wheels at any foot back to its neutral position.}
\label{fig:special_manoeuvres}
\end{figure}

The step which is considered by the planner at this level is an abstract step. We define an abstract step to be the direct transition from a pre-step pose to an after-step pose. It does not describe the motion sequence to perform the step and needs to be expanded to be stable and executable by the robot. An abstract step is visualized in~\cref{fig:special_manoeuvres}a. 

Each step is assigned a cost value
\begin{equation}
	C_S = k_7 \cdot L_{\text{step}} + k_8 \cdot (C_F(c_\text{h}) - 1) + k_9 \cdot \bigtriangleup H_{\text{step}} \text{,}
\label{eq:step_cost}
\end{equation}
where $k_7$ = 0.5, $k_8$ = 0.1 and $k_9$ = 2.3 which includes the step length $L_{\text{step}}$, the foot specific terrain difficulty costs of the foothold to step in $c_\text{h}$, and the maximum height difference $\bigtriangleup H_{\text{step}}$. If multiple end positions for a step exist, only the cheapest solution is returned to the search algorithm.

Further manoeuvres are required to navigate in cluttered environments. We define the footprint shown in~\cref{fig:step_criteria}a to be the neutral footprint. It provides high robot stability at a small footprint size, and is the preferred configuration for driving.

If both front feet are positioned in front of their neutral position, the robot may perform a longitudinal base shift manoeuvre. The base is shifted forward relatively to the feet until one of the front feet reaches its neutral position or a maximum leg length is reached for one of the rear legs (see~\cref{fig:special_manoeuvres}b). Base shifts of length $L_\text{BS}$ using the average discovered base costs $C_\text{B,avg}$ and $k_{10}$ = 0.5 carry the cost
\begin{equation}
	C_\text{BS} = k_{10} \cdot L_\text{BS} \cdot C_\text{B,avg} \text{.}
\end{equation}

If a rear foot is close to an obstacle, the pair of wheels at each front foot may be driven forward while keeping ground contact (\cref{fig:special_manoeuvres}c) which is a preparation for a rear foot step. If the robot footprint is not neutral, it may drive a single pair of wheels to their neutral position (\cref{fig:special_manoeuvres}d). A single foot movement of length $L_\text{FM}$ using the average discovered foot costs $C_\text{F,avg}$ and $k_{11}$ = 0.125 carries the cost
\begin{equation}
	C_\text{FM} = k_{11} \cdot L_\text{FM} \cdot C_\text{F,avg} \text{.}
\label{eq:foot_shift_cost}
\end{equation}

\begin{figure}
\centering
\includegraphics[width=0.5\linewidth]{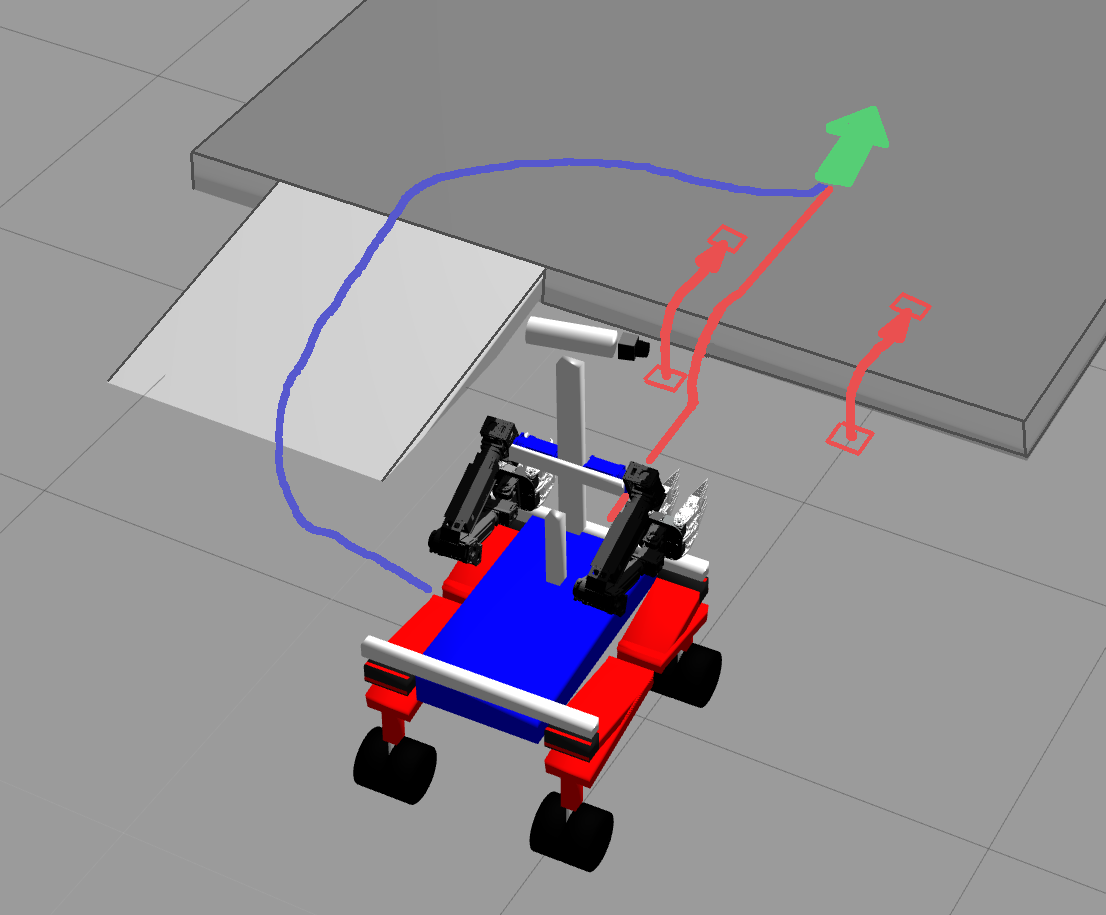}
\caption{Step-related manoeuvre costs are weighted such that the planner just prefers taking a 1.5\,m detour over a ramp (blue path) instead of stepping up (red path) to an elevated platform.}
\label{fig:momaro_ramp_edit}
\end{figure}

Since driving is faster and safer than stepping, we want the planner to consider drivable detours of acceptable length instead of including steps in the plan. We define that, when the robot stands in front of an 0.2\,m elevated platform, it should prefer a 1.5\,m long detour over a ramp instead of stepping up to this platform (\cref{fig:momaro_ramp_edit}). This can be achieved by increasing the costs of stepping-related manoeuvres by a certain factor.

\section{Step Motion Generation}

The result of the A* search is a cost-optimal abstract path which lacks executable motion sequences for steps and information about foot heights. The resulting path is expanded during step motion generation. It finds stable robot positions for steps and adds leg height information to the path. Again, costs are obtained from the cost client.

\subsection{Robot Stabilization}

\begin{figure}
\centering
\includegraphics[width=0.8\linewidth]{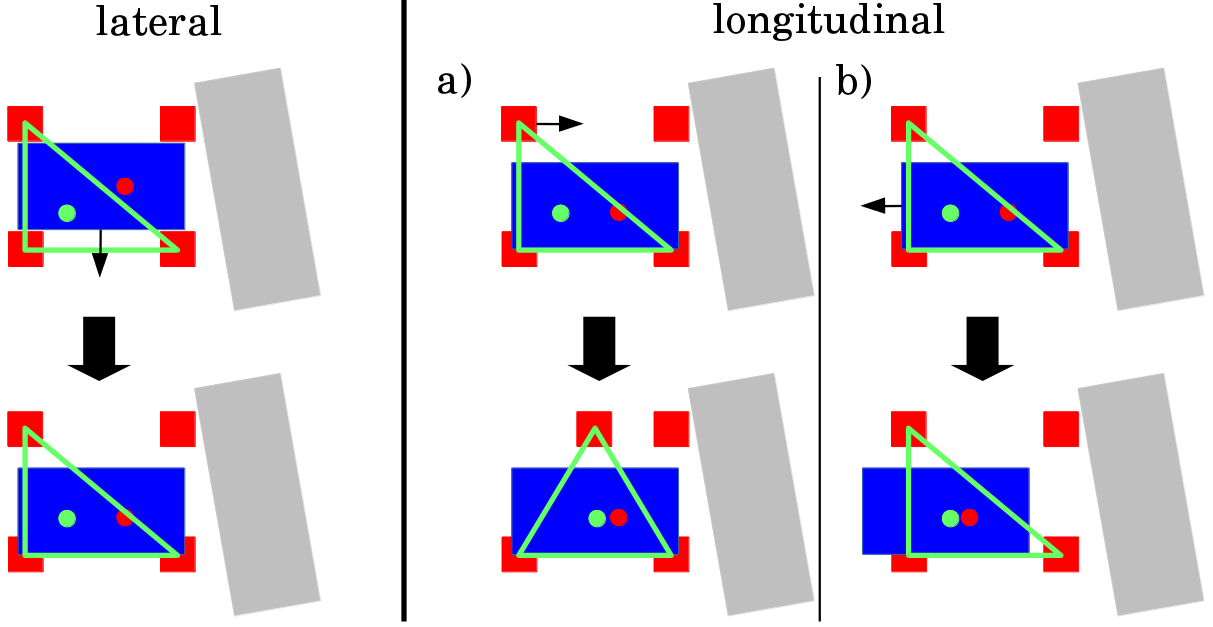}
\caption{To find a stable position for stepping, the robot CoM (red dot) needs to be aligned with the STC (green dot). Lateral alignment is done by base rolling. Longitudinal alignment is either done by rolling the remaining foot on the stepping side towards the center (a) or by shifting the robot base (b).}
\label{fig:step_generation}
\end{figure}

Abstract steps only describe the start and goal poses for a stepping manoeuvre. An executable transition between these poses is a motion sequence which keeps the robot stable at all times. Such a motion sequence is generated for each abstract step in the path. Due to the compliant leg design of our robot, we have no information about the exact position of the feet but have to estimate it from actuator feedback. Hence, we limit stability considerations to static stability. Since actuator speeds are slow, dynamic effects can be neglected. Stability estimation while stepping is done on the support triangle which is spanned by the horizontal position of the remaining three feet with ground contact (\cref{fig:step_generation}). If the horizontal robot center of mass (CoM) projection is inside the support triangle, the robot pose is stable. The closer the CoM is to the support triangle centroid (STC), the greater the stability. 

Lateral alignment of the CoM and STC is done by base roll motions (\cref{fig:step_generation}), which are rotations around the longitudinal axis. These are achieved by changing the leg lengths on one side of the robot. The resulting angle between the wheel axes and ground is compensated by the compliant legs and the soft-foam filled wheels. Roll manoeuvre parametrization is described in Sec.\,\ref{sec:leg_heights}.    

Longitudinal alignment of the CoM and STC is done by driving the remaining pair of wheels on the stepping side (e.g., the rear left foot if the front left foot is stepping) towards the robot center (\cref{fig:step_generation}). If this does not suffice because the motion is hindered by obstacles, the remaining longitudinal alignment is done by shifting the robot base. The longitudinal CoM position is also affected by the robot base pitch angle which is described in the next subsection. The presented motions generate a stable robot configuration which allows the desired step to be performed. After stepping, the robot reverses its base roll, foot displacement, and base shift manoeuvres to get back to its nominal configuration.

\subsection{Leg Heights}
\label{sec:leg_heights}

Each robot pose is assigned an individual height for each leg, which describes the vertical position of a foot relative to the robot base. When driving with neutral footprint, a low leg height of 0.27\,m is chosen, which provides a low CoM and good controllability through reduced leg compliance (see Fig.~\ref{fig:momaro} left). A larger ground clearance is chosen for manoeuvres other than driving to provide great freedom for leg movements (see Fig.~\ref{fig:momaro} right). In this case, the base height is determined by the highest foot. A soft constraint is applied that the leg height should be at least 0.45\,m. At the same time, a hard constraint from the mechanical system is that none of the legs exceeds its maximum leg length. The height of each individual foot can be read from the 2D height map. The robot base pitching angle is set to be 70\% of the ground slope, as can be seen in~\cref{fig:climbing_stairs}. This pitching value provides a good compromise between sufficient ground clearance for all four legs and a good CoM position.

\begin{figure}
\centering
\includegraphics[width=.9\linewidth]{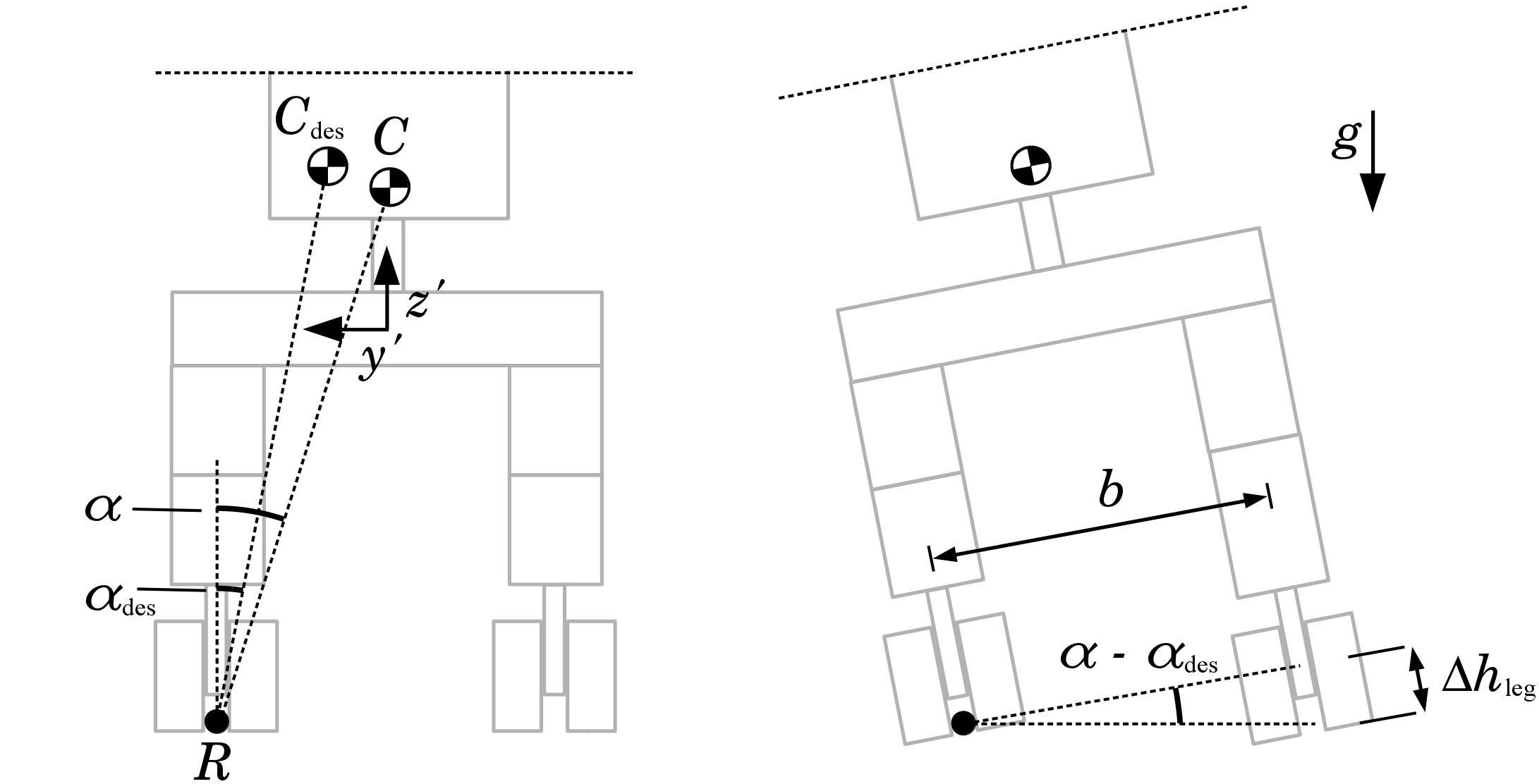}
\caption{Momaro's lower body in back view. Lateral CoM shifts can be achieved by changing leg length on one side which rolls the robot.}
\label{fig:roll_computation}
\end{figure}

\begin{figure*}
\centering
\includegraphics[width=\textwidth]{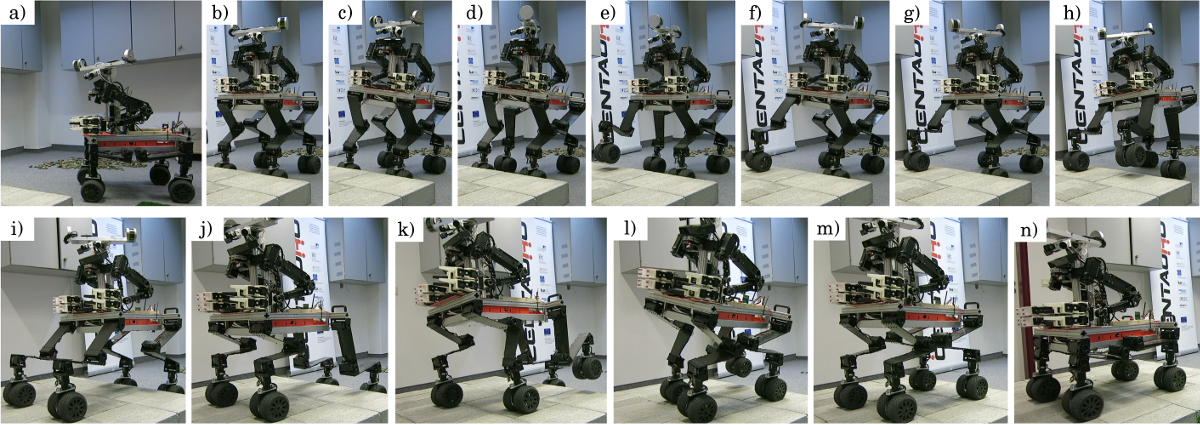}
\caption{Momaro stepping up an elevated platform. a) It arrives at the platform in low driving position, b) stands up, c) rolls its base to the left to shift its CoM laterally, d) drives its rear right pair of wheels forward to reach a stable stepping configuration. e) It then steps with its front right foot, f) drives its rear right pair of wheels back and g) rolls back its base to reach the configuration it had before the step. The remaining steps follow a similar motion sequence which is shown in less details. Subsequently, h) Momaro steps with its front left foot. i) It then drives forward and j) shifts its base forward before k) doing a step with the rear left and l) rear right foot. m) When the robot stands on the platform, n) it lowers its base to continue driving.}
\label{fig:real_stepping}
\end{figure*}

As described before, base roll manoeuvres are used to shift the robot CoM laterally. Due to the soft-foam filled wheels, we can estimate the center of rotation $R(y'_\text{rot}, z'_\text{rot})$ between the two wheels (\cref{fig:roll_computation}). In addition, the center of mass position $C(y'_\text{CoM}, z'_\text{CoM})$, the angle  
\begin{equation}
	\alpha = \text{arctan}\left(\frac{y'_\text{rot} - y'_\text{CoM}}{z'_\text{CoM} - z'_\text{rot}}\right)	
\end{equation}
between $\vec{RC}$ and the vertical axis and the desired lateral center of mass position $y'_\text{CoM,des}$ are given. Using \mbox{$\left\lVert \vec{RC} \right\rVert = \left\lVert \vec{RC}_\text{des} \right\rVert$} we compute the desired angle between $\vec{RC_\text{des}}$ and the vertical axis
\begin{equation}
	\alpha_\text{des} = \text{arcsin}\left( \frac{y'_\text{rot} - y'_\text{CoM,des}}{\left\lVert \vec{RC}\right\rVert}\right)
\end{equation}
and consequently using the footprint width $b$ we compute the desired leg height difference
\begin{equation}
	\bigtriangleup h_\text{leg} = b \cdot \text{tan}(\alpha - \alpha_\text{des})\text{.}
\end{equation}
This leg height difference is added to both legs on the respective side to induce a base roll manoeuvre. \cref{fig:real_stepping} shows how Momaro steps up an elevated platform, using the described motion sequences.

\section{Path Planning Extensions}

Due to the fine position and orientation resolution, the search space which is considered for path planning is large. Moreover, we want the planner to consider several detours before taking a step, which further increases planning times. We present methods to accelerate planning and to improve the resulting path quality. Their individual effects are investigated during evaluation.

\subsection{Robot Orientation Cost}

\begin{figure}
\centering
\includegraphics[width=\linewidth]{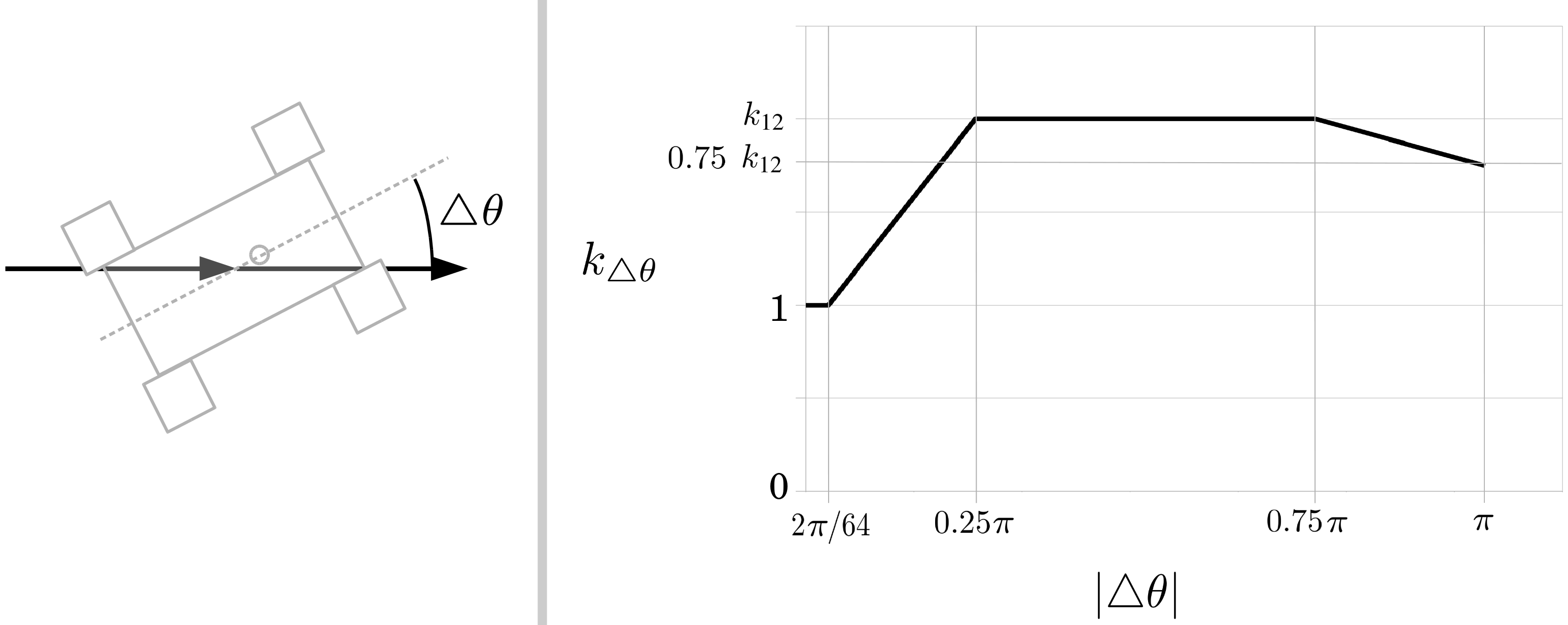}
\caption{For a difference between robot orientation and driving direction $\bigtriangleup \theta$ the cost factor $k_{\bigtriangleup \theta}$ is computed which expresses the desire to drive forward. It is 1 within an orientation step of $2\,\pi / 60$ and increases up to $k_\text{12}$. When driving backward, there is a desire to drive straight backwards since the required clearance is smaller.}
\label{fig:orientation_cost}
\end{figure}

Although our robot is capable of omnidirectional driving, there are multiple reasons to prefer driving forward. Since the sensor setup is not only used for navigation, but also for manipulation, it is designed to provide best measurement results for the area in front of the robot. The required width clearance is minimal when driving in a longitudinal direction, which is helpful in narrow sections such as doors. The same applies to driving backwards. Driving straight backwards requires a smaller clearance than driving diagonal backwards. Finally, our leg design restricts us to perform steps in the longitudinal direction. Thus, when approaching areas where stepping is required, a suitable orientation is helpful. We address this desire of preferring special orientations by multiplying neighbour costs during A* search by the individual factor $k_{\bigtriangleup \theta}$, as described in~\cref{fig:orientation_cost}.

\subsection{ARA*}

To obtain feasible paths quickly, we extend the A* algorithm to an Anytime Repairing A* (ARA*)~\cite{likhachev2003ara}. Its initial search provides solutions with bounded suboptimality by giving the heuristic a weight \textgreater\,1. The result quality is then improved by decreasing the heuristic weight stepwise down to 1, if the given planning time is not exhausted yet. ARA* recycles the representations it generated previously to accelerate replanning.

\begin{figure}
\centering
\includegraphics[width=\linewidth]{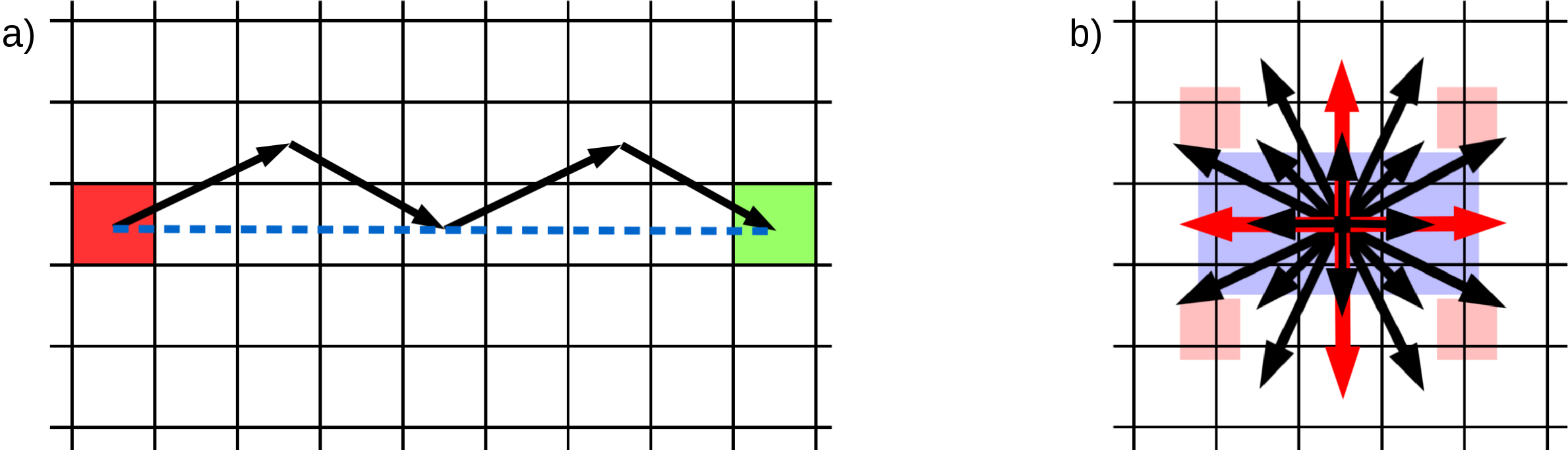}
\caption{Adressing ARA* preferences of long moves. a) For larger heuristic weights, the ARA* algorithm prefers those driving manoeuvres which bring the robot as close to the goal as possible which leads to undesired paths (black arrows). b) To obtain the desired behaviour (blue line) we extend the driving neighbourhood by the four red manoeuvres.}
\label{fig:ara_change_neighbourhood}
\end{figure}

When planning with higher weighted heuristics, the planner prefers those driving manoeuvres that bring it as close to the goal as possible. This leads to the undesired effect that resulting paths mainly consist of those driving manoeuvres which go two cells in one direction and one cell in an orthogonal direction, as can be seen in Fig.~\ref{fig:ara_change_neighbourhood}~a. To prevent this behaviour, we extend the driving neighbourhood size from 16 to 20, as shown in~\cref{fig:ara_change_neighbourhood}~b.

\section{Plan Execution}

We utilize the control framework described by Schwarz et al.~\cite{nimbro_rescue}. Input for omnidirectional driving is a velocity command $\vec{w} = (v_{x'}, v_{y'}, \omega)$ with horizontal linear velocities $v_{x'}$ and $v_{y'}$ in robot coordinates and a rotational velocity $\omega$ around the vertical robot axis. We obtain $\vec{w}$ by computing a B-spline through the next five driving poses and aim towards a pose $\vec{p} = (p_x, p_y, p_\theta)$ on this B-spline in front of the robot. 

For manoeuvres which require leg movement, the input to the control framework are 2D $(x', z')$ foot poses which can be directly derived from the resulting path.

\section{Experiments}

We evaluate our path planning method and the presented extensions in the Gazebo simulation environment\footnote{\url{http://www.gazebosim.org}}. Experiments are done on one core of a 2.6\,GHz Intel i7-6700HQ processor using 16\,GB of memory. A video of the experiments is available online\footnote{\url{https://www.ais.uni-bonn.de/videos/IROS_2017_Klamt/}}.  

\begin{figure}
\centering
\includegraphics[width=0.7\linewidth]{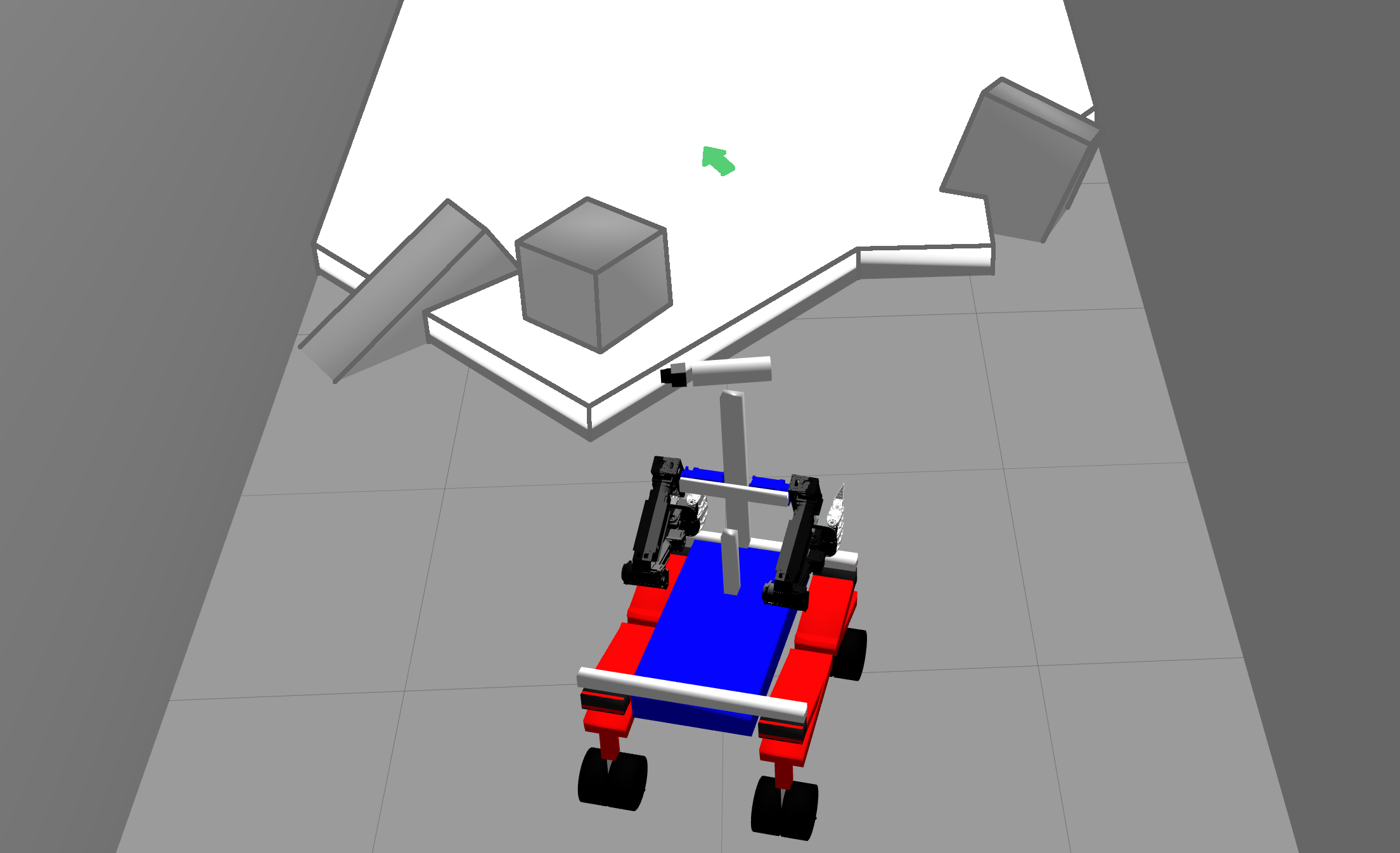}
\caption{Gazebo scenario to compare planner variants. Momaro stands in front of an elevated platform, cluttered with obstacles and has to reach a pose on this platform.}
\label{fig:hybrid_scenario}
\end{figure}

\begin{figure}
\centering
\includegraphics[width=.75\linewidth]{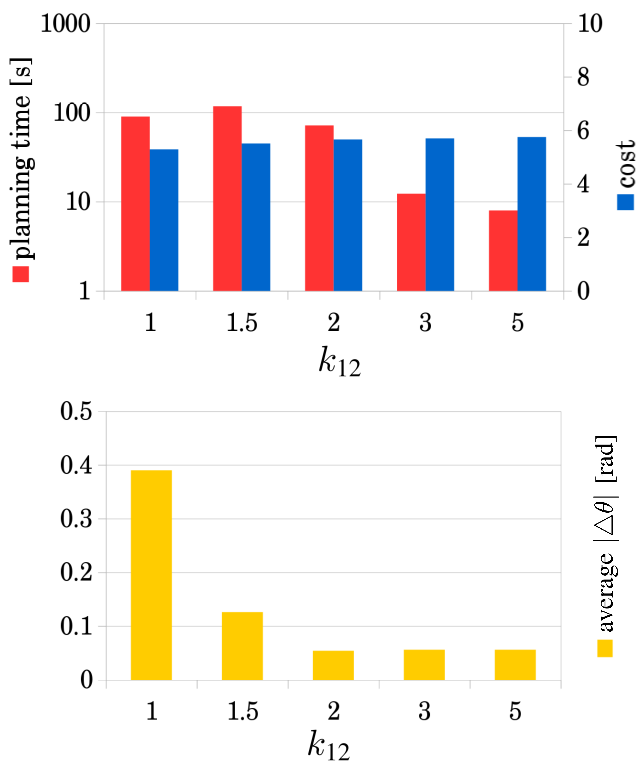}
\caption{Comparison between the original A* planner ($k_\text{12} = 1$) and the modification with robot orientation cost factor.}
\label{fig:acceleration_chart}
\end{figure}

\begin{figure}
\centering
\includegraphics[width=\linewidth]{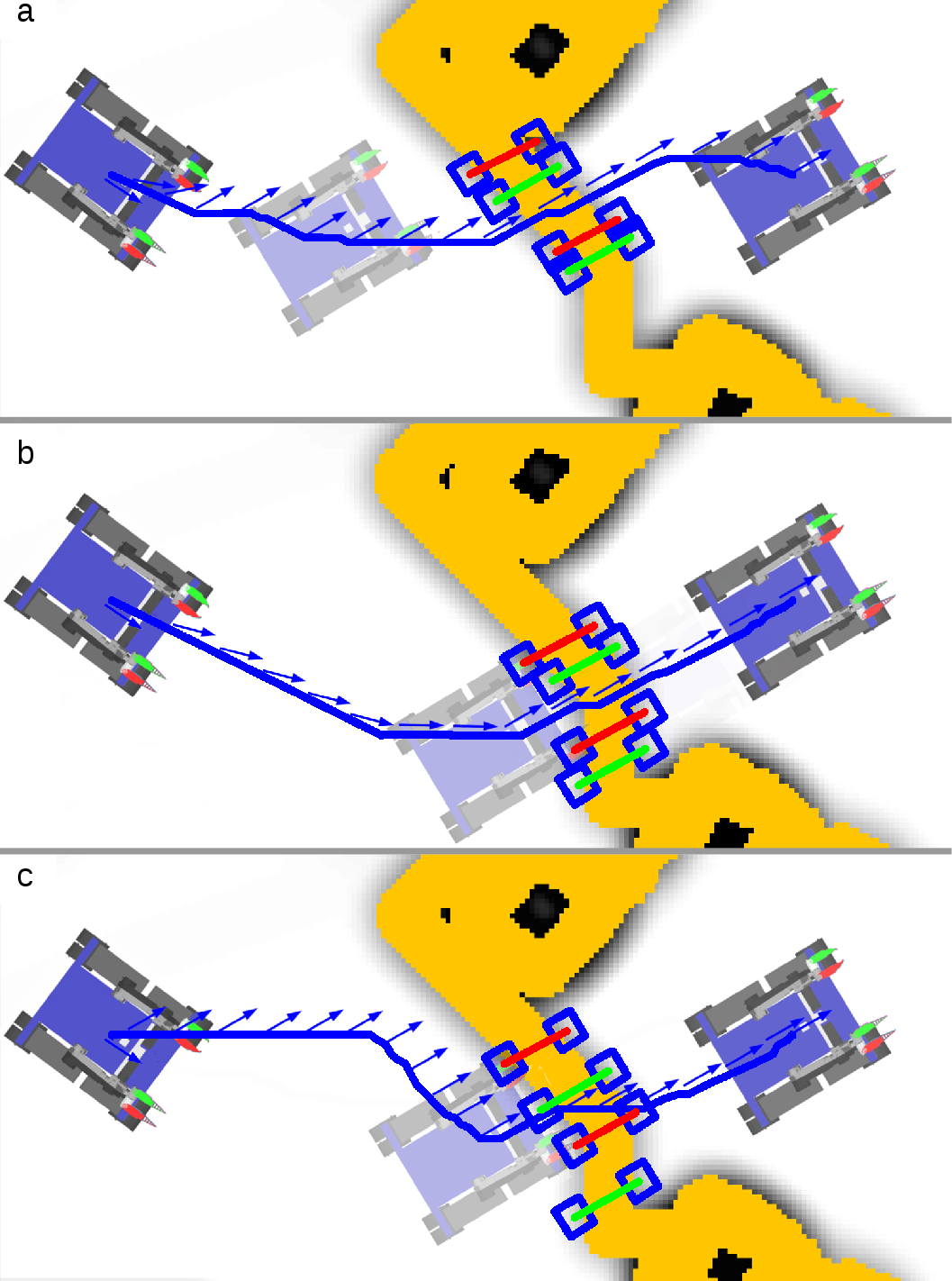}
\caption{Resulting paths of our planner in different settings on a foot cost map. Yellow areas are not traversable by driving. Blue paths show the robot center position, arrows show the orientation. Blue rectangles show used footholds. Red lines represent front foot steps, green lines represent rear foot steps. a) Result of the plain A* algorithm, b) orientation differences are considered ($k_\text{12} = 2$), c) first result of the ARA* algorithm using a heuristic weight of 3.0.}
\label{fig:ara_paths}
\end{figure}

In a first scenario, the robot stands in a corridor in front of an  elevated platform and some cluttered obstacles. It needs to find a way to a goal pose on this platform as can be seen in~\cref{fig:hybrid_scenario}. We compare the performance of the planner for different values of $k_\text{12}$ in~\cref{fig:acceleration_chart}. The parameter $k_\text{12}$ is defined in~\cref{fig:orientation_cost}. All shown path costs are the costs the path would have in the plain A* planner to keep them comparable. It can be seen that an increasing robot orientation cost factor decreases the difference between robot orientation and driving direction. This can also be observed in~\cref{fig:ara_paths}. Moreover, planning is accelerated for higher values ($\geq 2$) of $k_\text{12}$, while path costs increase only slightly.

\begin{figure}
\centering
\includegraphics[width=.9\linewidth]{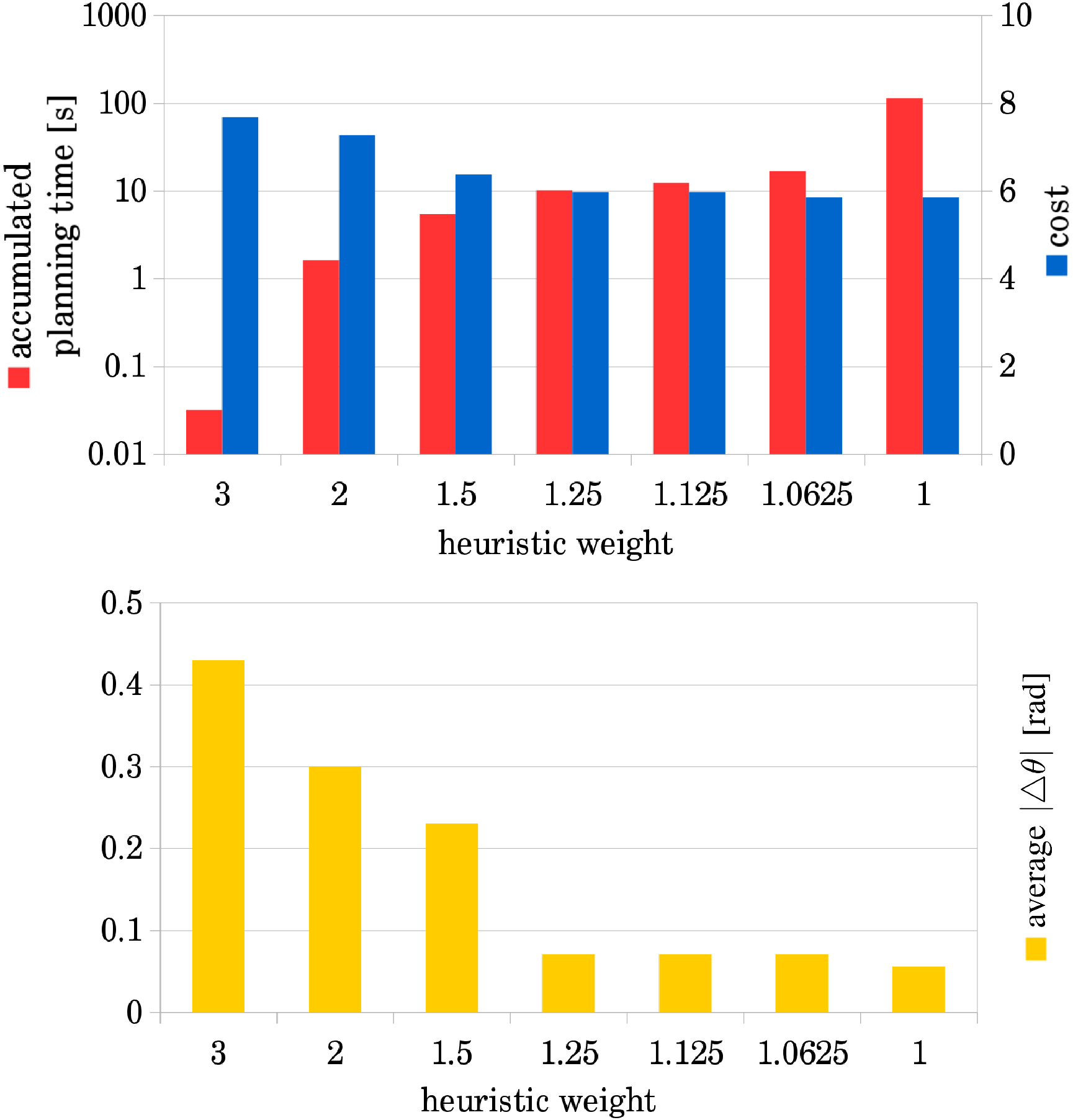}
\caption{Performance and result quality of the ARA* algorithm where $k_\text{12} = 2$.}
\label{fig:ara_chart}
\end{figure}

In addition, we evaluate the ARA* approach in the same scenario. We choose exponentially decaying heuristic weights, starting at 3.0 while $k_\text{12}$ = 2. The performance results are shown in~\cref{fig:ara_chart} and a path can be seen in~\cref{fig:ara_paths}. ARA* provides its first result in 32\,ms, and this is 31\% more costly than the optimal solution. A solution with only 2\% higher costs is found in \texttildelow 10\,s which is sufficiently fast compared to the required execution times. Planning paths using an optimal heuristic weight takes infeasibly long (\textgreater\,$100\,s$). Searching optimal results takes longer than in the plain A* variant because higher heuristic weights are considered first and the neighbourhood size changed from 16 to 20. It can be seen that the effect of the robot orientation cost factor increases with decreasing heuristic weights.

\begin{figure}
\centering
\includegraphics[width=0.75\linewidth]{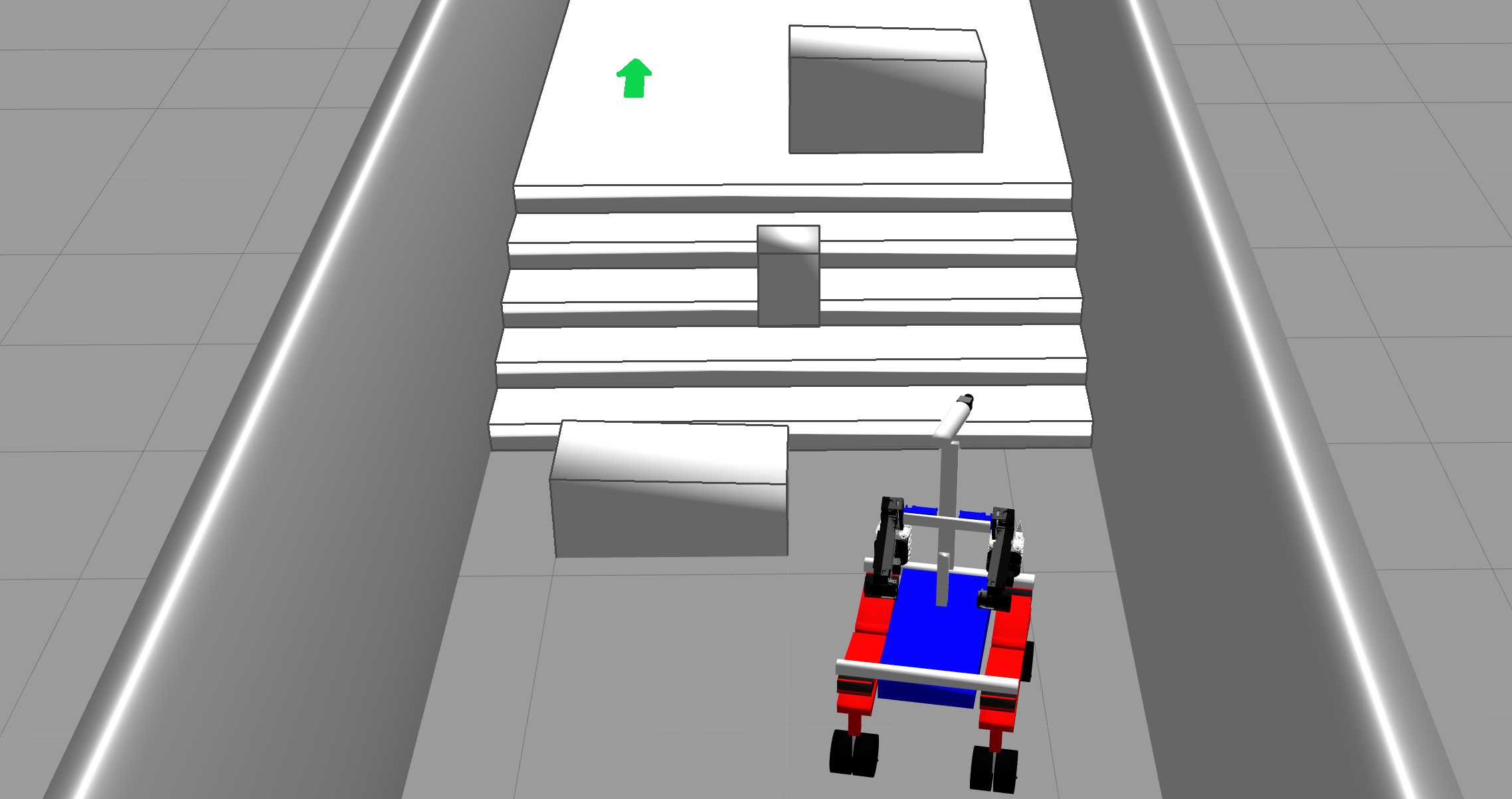}
\caption{Challenging scenario to demonstrate the planner capabilities. A staircase with obstacles on it requires a combination of stepping and driving sideways.}
\label{fig:stair_scenario}
\end{figure}

\begin{figure}[h]
\centering
\includegraphics[width=0.6\linewidth]{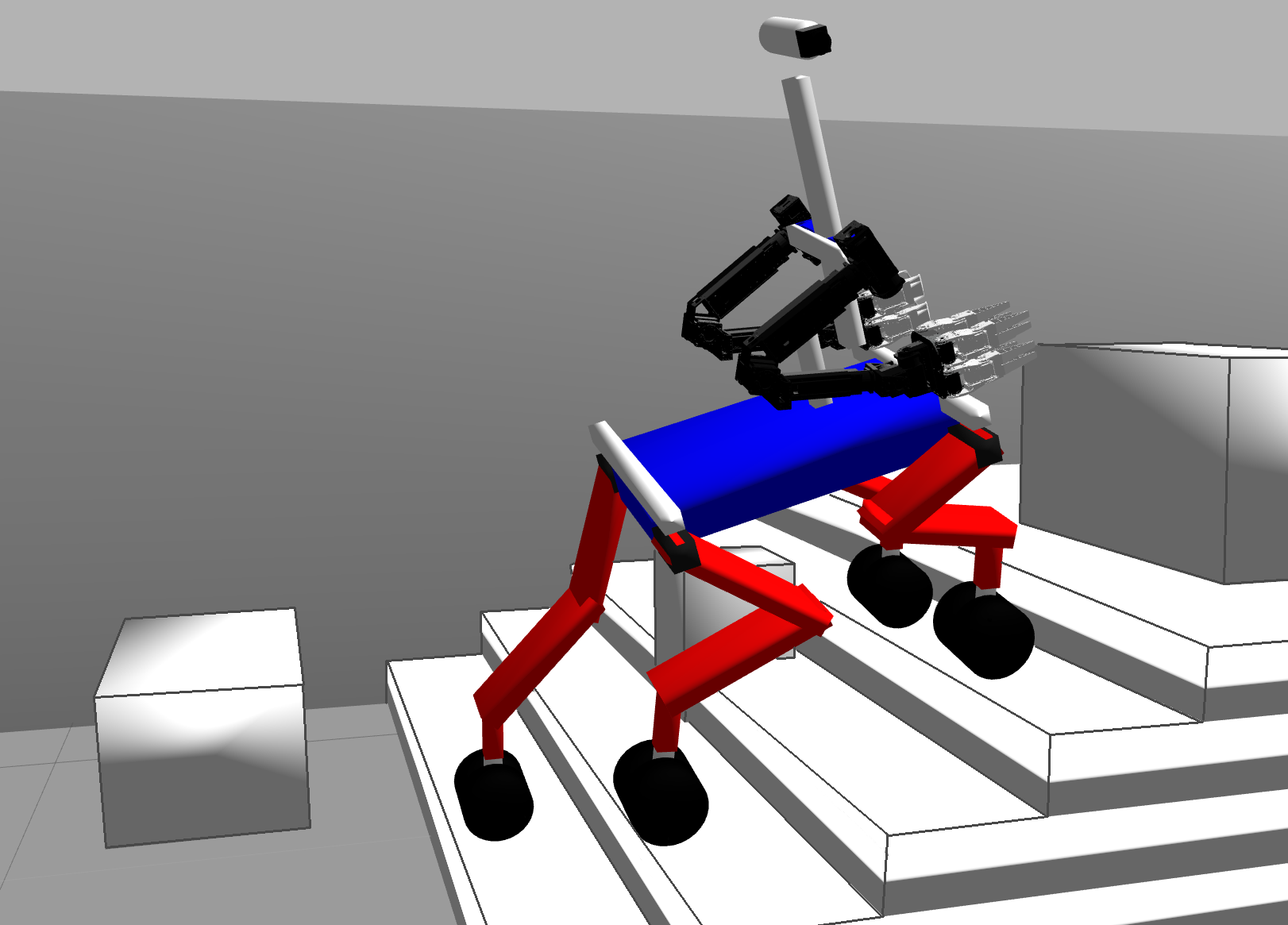}
\caption{Momaro climbing stairs. The robot base pitch angle adapts to 70\% of the terrain slope.}
\label{fig:climbing_stairs}
\end{figure}

To demonstrate the capabilities of our planner, we present a second experiment in which Momaro has to climb a staircase which is blocked by obstacles (\cref{fig:stair_scenario}). Tracked vehicles would have great difficulties to overcome this. Our robot climbs the stairs and then drives sideways while taking the obstacle between its legs. Our planner finds a first path with heuristic weight of 3.0 in 1.02\,s. \cref{fig:climbing_stairs} shows Momaro on the staircase and visualizes how the robot base adapts its pitch angle to the terrain slope.

\begin{figure}
\centering
\includegraphics[width=0.85\linewidth]{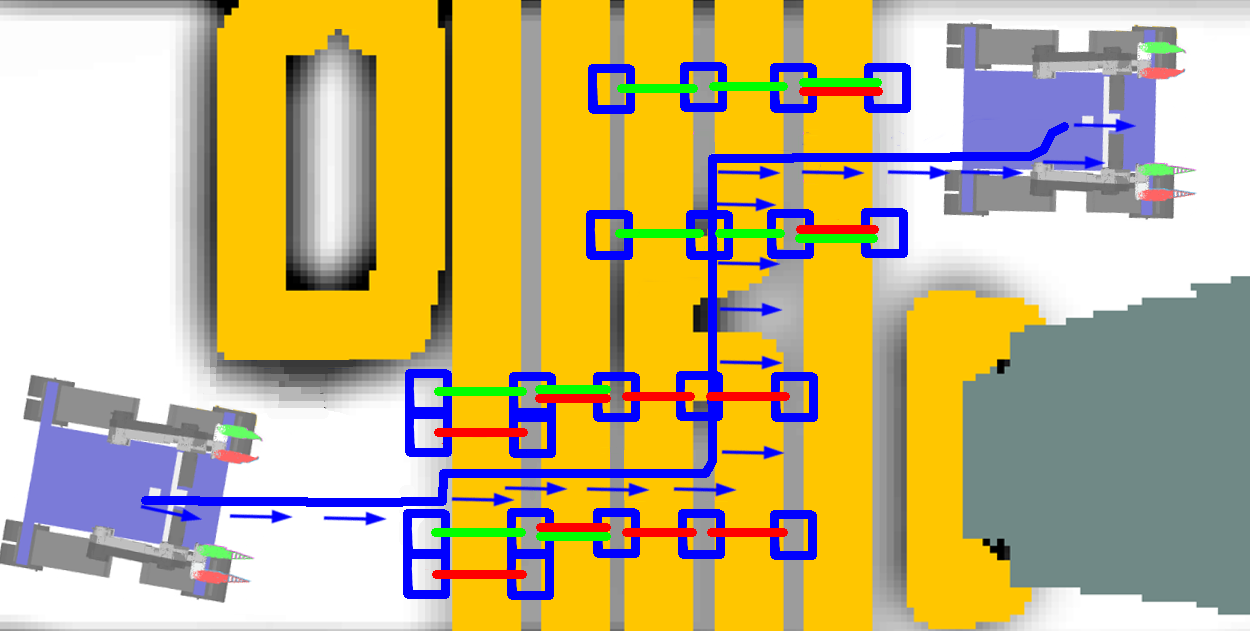}
\caption{Planner output for the staircase scenario on a foot cost map using a heuristic weight of 3. The blue path shows the robot center position. Arrows show the robot orientation. Blue squares show used footholds. Red lines represent front foot steps; green lines represent rear foot steps.}
\label{fig:stair_path}
\end{figure}

\section{Conclusion}

In this paper, we presented a hybrid locomotion planning approach which combines driving and stepping in a single planner. It provides paths with bounded suboptimality in feasible time and is capable of path planning in challenging environments. Due to the high dimensionality of the search space and the desire to consider detours instead of stepping, finding optimal solutions may take considerable time. We address this by using an anytime approach with larger heuristic weights. 
The planned paths are executed by our mobile manipulation robot Momaro.
Experiments demonstrated that our method generates paths for challenging terrain, which could not be traversed by driving or stepping alone.



\bibliographystyle{IEEEtranN}
\bibliography{references}

\addtolength{\textheight}{-12cm}   





\end{document}